\documentclass[conference,a4paper]{IEEEtran}
\IEEEoverridecommandlockouts

\usepackage{cite}
\usepackage{amsmath,amssymb,amsfonts}
\usepackage{graphicx}
\usepackage{textcomp}
\usepackage{xcolor}
\usepackage{booktabs}
\usepackage{array}
\usepackage{balance}
\usepackage{url}

\def\BibTeX{{\rm B\kern-.05em{\sc i\kern-.025em b}\kern-.08em
    T\kern-.1667em\lower.7ex\hbox{E}\kern-.125emX}}

\begin{document}

\title{Validation-Stage Combinatorial Fusion Analysis for Imbalanced Credit-Card Fraud Detection}

\author{%
\IEEEauthorblockN{Xiao Han\IEEEauthorrefmark{1}}
\IEEEauthorblockA{\textit{Goizueta Business School}\\
\textit{Emory University}\\
Atlanta, Georgia, USA\\
xhan@alumni.emory.edu}
\and
\IEEEauthorblockN{Chenyu Wu\IEEEauthorrefmark{1}}
\IEEEauthorblockA{\textit{Pratt School of Engineering}\\
\textit{Duke University}\\
Durham, North Carolina, USA\\
wuchenyu999@outlook.com}
\thanks{\IEEEauthorrefmark{1}Xiao Han and Chenyu Wu contributed equally to this work. Corresponding author: Xiao Han.}
}

\maketitle

\begin{abstract}
Credit-card fraud detection is difficult because fraudulent transactions are rare, costly, and unevenly distributed. Strong gradient-boosted tree models already perform well on structured transaction data, so the value of another fusion method is not obvious. This paper examines whether Combinatorial Fusion Analysis (CFA), which searches over model subsets and rank-score fusion rules, can still add value on the IEEE-CIS Fraud Detection benchmark. Using a leakage-free 60/20/20 train/validation/test protocol, we evaluate 480 fusion configurations built from seven base classifiers. The best test-set result comes from diversity-weighted score fusion of Random Forest, XGBoost, and LightGBM (DEF WtScore), with AUC-ROC = 0.9405, AUPRC = 0.6699, and F1 = 0.6373. Bootstrap confidence intervals from 1,000 resamples show that the gains over the strongest single model exclude zero for all three metrics. CFA matches soft voting on AUC-ROC, improves AUPRC and F1, and outperforms stacking in this setting. A CTGAN augmentation experiment gives a negative result: synthetic fraud samples degrade both individual models and CFA. Overall, CFA is most useful here not as a way to combine every classifier, but as a validation-stage method for choosing a small, complementary subset and assigning diversity-aware weights.
\end{abstract}


\section{Introduction}
Payment fraud is a demanding setting for machine learning. Fraudulent transactions are rare, the cost of mistakes is uneven, and behavior can change over time. A false alert may block a legitimate customer or create manual-review cost; a missed fraud can create direct financial loss. For that reason, a useful model should not only classify the majority class correctly but also rank high-risk transactions well and support a defensible operating threshold.

The IEEE-CIS Fraud Detection benchmark (Kaggle, 2019) is a useful public testbed for this problem. It contains 590,540 anonymized transactions with mixed feature types and substantial missingness. Although it is not a live production deployment dataset, its transaction structure, class imbalance, and missing-data patterns make it valuable for comparing fraud-detection methods. On this dataset, boosted-tree models such as XGBoost and LightGBM are hard baselines. In preliminary experiments and in prior Kaggle-style studies, simple ensembles rarely improve them by much. The question is therefore not whether fusion can rescue weak models. The question is whether a principled fusion method can produce a reliable marginal gain after strong structured-data models have already been tuned.

We study that question using Combinatorial Fusion Analysis (CFA). CFA represents each classifier through its Rank-Score Characteristic (RSC) function, measures diversity between scoring systems, and searches both rank-based and score-based fusion rules across model subsets. In this paper, we use CFA as a validation-stage framework for subset and fusion-rule selection. Two questions guide the experiments. RQ1 asks whether this validation-stage search improves over the best single model and over standard voting and stacking, or whether it mainly recovers uniform voting on a good subset. RQ2 asks which subsets and fusion rules are selected, and whether the gain comes from useful disagreement among strong learners rather than from adding weaker but more unusual models. We answer both questions under a strict train/validation/test design: preprocessing and base models are fit on the training set; thresholds, subsets, and fusion rules are selected on the validation set; the test set is used only for final reporting.

The paper contributes four empirical findings. First, it evaluates CFA on IEEE-CIS with seven base learners, including linear models, random forest, XGBoost, LightGBM, and TabNet, under a protocol designed to avoid leakage. Second, it shows that the selected ensemble is a compact tree-based subset, not the full model library and not the most diverse set of models. This matters because diversity helps only when the added model contributes useful errors. Third, it uses bootstrap confidence intervals to separate a small but reliable gain from search noise. Fourth, it reports a bounded negative result on CTGAN augmentation, which is useful for practitioners working with high-dimensional, mixed-type fraud data.

\section{Literature Review}

\subsection{Fraud Detection and Ensemble Learning}
Credit-card fraud detection has long been used to study imbalanced classification. Bhattacharyya \textit{et al.} compared logistic regression, support-vector machines, and random forests, and found that nonlinear ensemble methods were generally stronger on card-transaction data \cite{bhattacharyya2011}. Dal Pozzolo \textit{et al.} discuss practical issues that make fraud modeling different from ordinary binary classification, including sampling bias, delayed labels, and changing fraud patterns \cite{dalpozzolo2014,dalpozzolo2018}. Carcillo \textit{et al.} show that unsupervised anomaly scores and supervised classifiers can complement each other in large-scale fraud systems \cite{carcillo2021}. These studies point to the same lesson that accuracy of a single and how models fail both matter.

Evaluation is also more delicate when labels are highly skewed. He and Garcia review learning from imbalanced data \cite{he2009}, Fawcett gives the standard treatment of ROC analysis \cite{fawcett2006}, and Saito and Rehmsmeier explain why precision-recall curves can be more informative when the positive class is rare \cite{saito2015}. These evaluation concerns motivate our use of both AUC-ROC and AUPRC, along with threshold-level metrics.

Boosted decision trees remain strong baselines for structured fraud data. XGBoost \cite{chen2016} and LightGBM \cite{ke2017} model nonlinear interactions efficiently and handle heterogeneous feature distributions well. Recent studies also use stacking. El Bazi \textit{et al.} combine boosted-tree models with Bayesian optimization for credit-card fraud detection \cite{elbazi2024}. Diversity-aware aggregation is relevant in non-stationary fraud settings as well. Paldino \textit{et al.} show that ensemble diversity can improve fraud detection under recurrent concept drift \cite{paldino2024}. Our study differs by making both the subset and the fusion rule validation-selected choices rather than fixed design decisions.

\subsection{Deep Models and Synthetic Augmentation}
Deep learning has had mixed results on structured fraud benchmarks. TabNet uses sequential attention over features and offers an interpretable deep architecture for structured data \cite{arik2021}. Sequence models can use cardholder histories when temporal behavioral data are available \cite{jurgovsky2018}, but the IEEE-CIS release is a flat transaction-level benchmark. Therefore, we use TabNet as a deep structured-data learner, not as a sequential model.

Class imbalance is commonly addressed through weighting, sampling, or synthetic generation. SMOTE interpolates minority-class observations in feature space \cite{chawla2002}, and undersampling can distort posterior probabilities unless calibration is handled carefully \cite{dalpozzolo2015}. CTGAN models mixed-type data through conditional generation \cite{xu2019}, and GAN-based oversampling has been applied to credit-card fraud detection \cite{fiore2019}. Whether such augmentation helps on a high-dimensional, label-encoded fraud benchmark is an empirical matter. Therefore, we report the CTGAN experiment separately rather than assuming that synthetic fraud samples will help.

\subsection{Combinatorial Fusion Analysis}
CFA was introduced to combine multiple scoring systems through their Rank-Score Characteristic functions \cite{hsu2006}. The RSC function describes how a model distributes scores across ranked instances. Two models can have similar aggregate accuracy but different score profiles. Hsu and Taksa show that rank and score combination are not interchangeable, and their relative performance depends on the score shape and diversity of the systems being fused \cite{hsu2005}. The RSC-based cognitive-diversity measure provides the basis for diversity-weighted fusion \cite{hsu2010}. CFA has been used in information retrieval, ChIP-seq peak detection, and recent multi-model deep-learning fusion \cite{schweikert2012,lin2024,hsu2024}, while it has not previously been evaluated on IEEE-CIS with both TabNet and CTGAN included in the design.

\section{Methodology}

\subsection{Problem Formulation}
Let $D=\{(x_i,y_i)\}_{i=1}^{n}$ denote the IEEE-CIS dataset, where $x_i$ is a feature vector and $y_i\in\{0,1\}$ indicates fraud. Each base classifier $m$ outputs a fraud score $s_m(x)$. Given a candidate model subset $S$ and a fusion rule $f$, the goal is to select $(S^*,f^*)$ on the validation set and then report performance on a held-out test set.

\subsection{Evaluation Protocol}
We use a 60/20/20 split. All preprocessing steps are fit only on the training data and then applied to validation and test data. The validation set is used for three choices: the F1 threshold, the CFA model subset, and the fusion rule. The test set is used once, after those choices have been fixed.

\begin{table}[t]
\caption{Evaluation splits.}
\label{tab:splits}
\centering
\scriptsize
\begin{tabular}{lrrp{0.27\linewidth}}
\toprule
Split & Samples & Frauds & Purpose \\
\midrule
Train (60\%) & 354,324 & 12,398 & fit models and preprocessing \\
Validation (20\%) & 118,108 & 4,133 & select CFA rule and threshold \\
Test (20\%) & 118,108 & 4,132 & final reporting only \\
\bottomrule
\end{tabular}
\end{table}

\subsection{Base Classifiers}
The model library contains seven classifiers: instance-based, linear, bagging, boosting, and attention-based learners. After dropping 214 columns with more than 50\% missing values, 218 predictive features remain. KNN and LDA are trained on PCA-50 because of dimensionality and scaling sensitivity; the tree ensembles and TabNet use the full feature set.

\begin{table}[t]
\caption{Base classifiers and feature spaces.}
\label{tab:models}
\centering
\scriptsize
\begin{tabular}{llll}
\toprule
Label & Model & Type & Features \\
\midrule
A & KNN & Instance-based & PCA-50 \\
B & LDA & Generative linear & PCA-50 \\
C & Logistic Regression & Discriminative linear & Full (218) \\
D & Random Forest & Bagging & Full (218) \\
E & XGBoost & Boosting & Full (218) \\
F & LightGBM & Boosting & Full (218) \\
G & TabNet & Deep tabular & Full (218) \\
\bottomrule
\end{tabular}
\end{table}

\subsection{RSC Diversity and Fusion Rules}
For each model $m$, samples are sorted by decreasing $s_m(x_i)$. The RSC function maps normalized rank positions to model scores. We compute diversity between two systems as the average $L_1$ distance between their RSC functions:
\begin{equation}
 d(i,j)=\frac{1}{n}\sum_k\left|RSC_i(k)-RSC_j(k)\right|.
\end{equation}
The diversity strength of model $i$ inside subset $S$ is
\begin{equation}
 DS(i,S)=\sum_{j\in S,\,j\ne i} d(i,j).
\end{equation}
For every subset $S$ with $|S|\geq2$, we evaluate four fusion rules. Average Score and Weighted Score are
\begin{equation}
\begin{aligned}
 f_{AS}(x)&=\frac{1}{|S|}\sum_{m\in S}s_m(x),\\
 f_{WS}(x)&=\frac{\sum_{m\in S}DS(m,S)s_m(x)}{\sum_{m\in S}DS(m,S)}.
\end{aligned}
\end{equation}
Average Rank and Weighted Rank apply the same idea to ranks. With seven base models, the search covers 120 model subsets and 480 fusion cases.

\subsection{Preprocessing, Tuning, and Inference}
The transaction and identity tables are merged on \texttt{TransactionID}. Columns with more than 50\% missingness are dropped using a threshold estimated on the training set only. Categorical features are label-encoded on the training set. Unseen validation or test categories are mapped to a sentinel value. Numeric missing values are imputed with training medians. \texttt{StandardScaler} and PCA-50 are fit only on training data for the models that require them.

All models except LDA are tuned by stratified 5-fold cross-validation on the training set, optimizing AUPRC. Random forest uses balanced class weights. XGBoost and LightGBM search \texttt{scale\_pos\_weight} in \{1, 5, 10, 20\}. TabNet uses early stopping on validation AUC with patience 15. For inference, we compute 95\% bootstrap confidence intervals for the test-set difference between CFA and LightGBM using 1,000 resamples. Resamples containing only one class are discarded.

\section{Experimental Setup}
The merged IEEE-CIS table contains 434 columns, including identifiers and the \texttt{isFraud} label. The dataset has 20,663 fraudulent transactions, a 3.50\% fraud rate. Identity information is available for about one quarter of transactions. The stratified split preserves this class balance across train, validation, and test.

\begin{figure*}[t]
\centering
\includegraphics[width=0.66\textwidth]{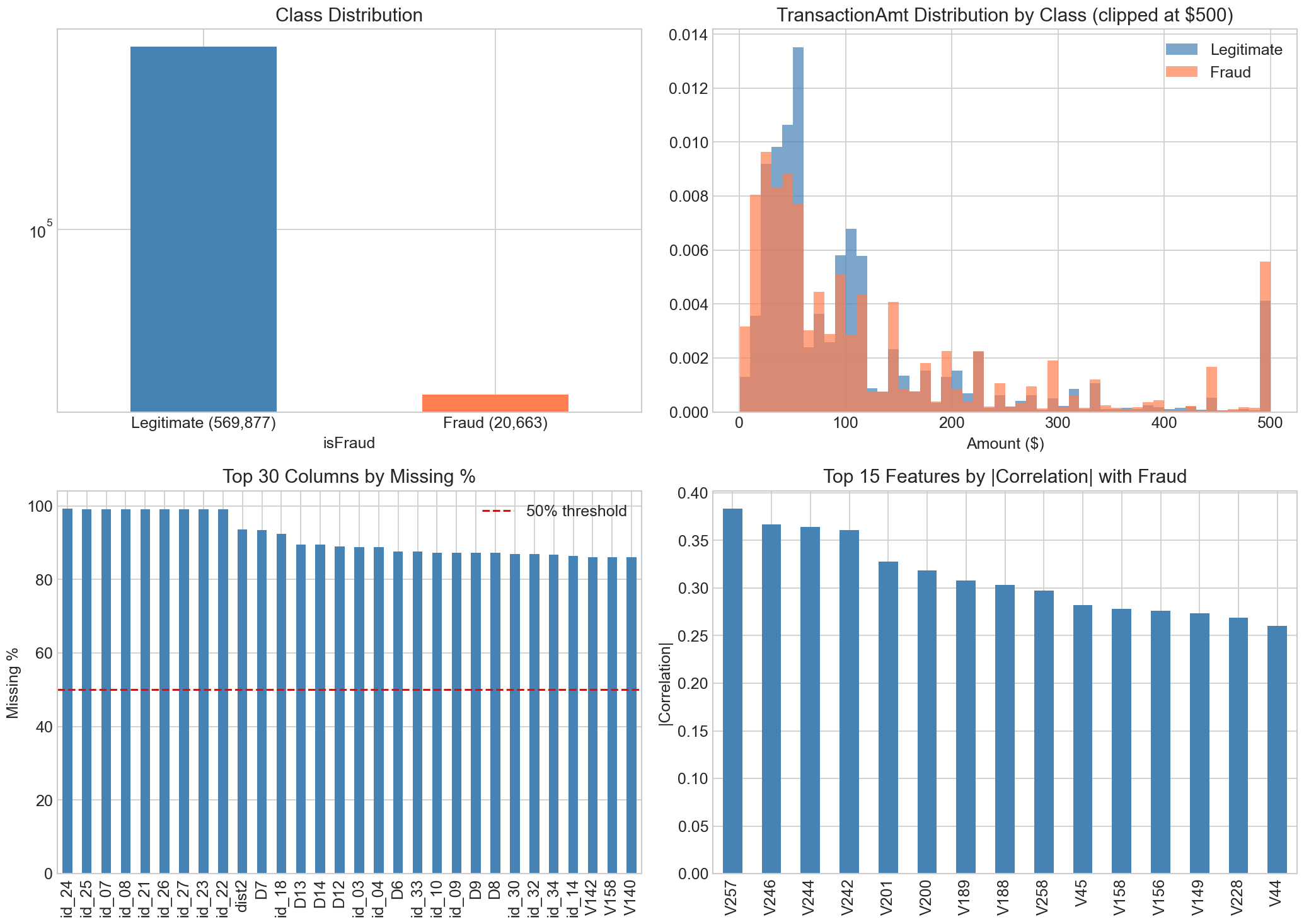}
\caption{Exploratory structure of the IEEE-CIS data: class imbalance on a log scale, transaction amount distribution by class, high-missingness columns with the 50\% drop threshold, and the strongest absolute feature correlations with fraud.}
\label{fig:eda}
\end{figure*}

We report AUC-ROC for overall ranking and AUPRC for retrieval of the rare fraud class. For threshold-level behavior, we report F1 at a validation-selected threshold together with precision and recall. F1 is not treated as the only operational objective. In production, fraud teams often care about precision at a fixed alert budget, recall at a fixed false-positive rate, or expected cost under a chosen loss ratio. We include additional operating-point metrics for this reason.

\section{Results}

\subsection{Base Model Performance}
Table~\ref{tab:base} reports held-out test performance. LightGBM has the highest single-model AUC-ROC and AUPRC, while XGBoost has the highest single-model F1. The tree models clearly outperform the linear models, which suggests that the useful signal in IEEE-CIS is strongly nonlinear. TabNet performs better than the linear baselines but does not match the tuned boosted trees.

\begin{table}[t]
\caption{Base model performance on the test set.}
\label{tab:base}
\centering
\scriptsize
\begin{tabular}{lrrrrrr}
\toprule
Model & AUC & AUPRC & F1 & Prec. & Rec. & Thr. \\
\midrule
A KNN & 0.8081 & 0.4127 & 0.4401 & 0.5737 & 0.3570 & 0.2719 \\
B LDA & 0.8027 & 0.2127 & 0.2683 & 0.2265 & 0.3291 & 0.1903 \\
C LR & 0.8327 & 0.3470 & 0.3802 & 0.4006 & 0.3618 & 0.2058 \\
D RF & 0.9048 & 0.5982 & 0.5769 & 0.6177 & 0.5411 & 0.2050 \\
E XGB & 0.9335 & 0.6402 & 0.6126 & 0.6692 & 0.5649 & 0.6564 \\
F LGBM & 0.9343 & 0.6480 & 0.6093 & 0.6640 & 0.5629 & 0.6503 \\
G TabNet & 0.8717 & 0.4161 & 0.4631 & 0.4994 & 0.4318 & 0.9420 \\
\bottomrule
\end{tabular}
\end{table}

\begin{figure}[t]
\centering
\includegraphics[width=\linewidth]{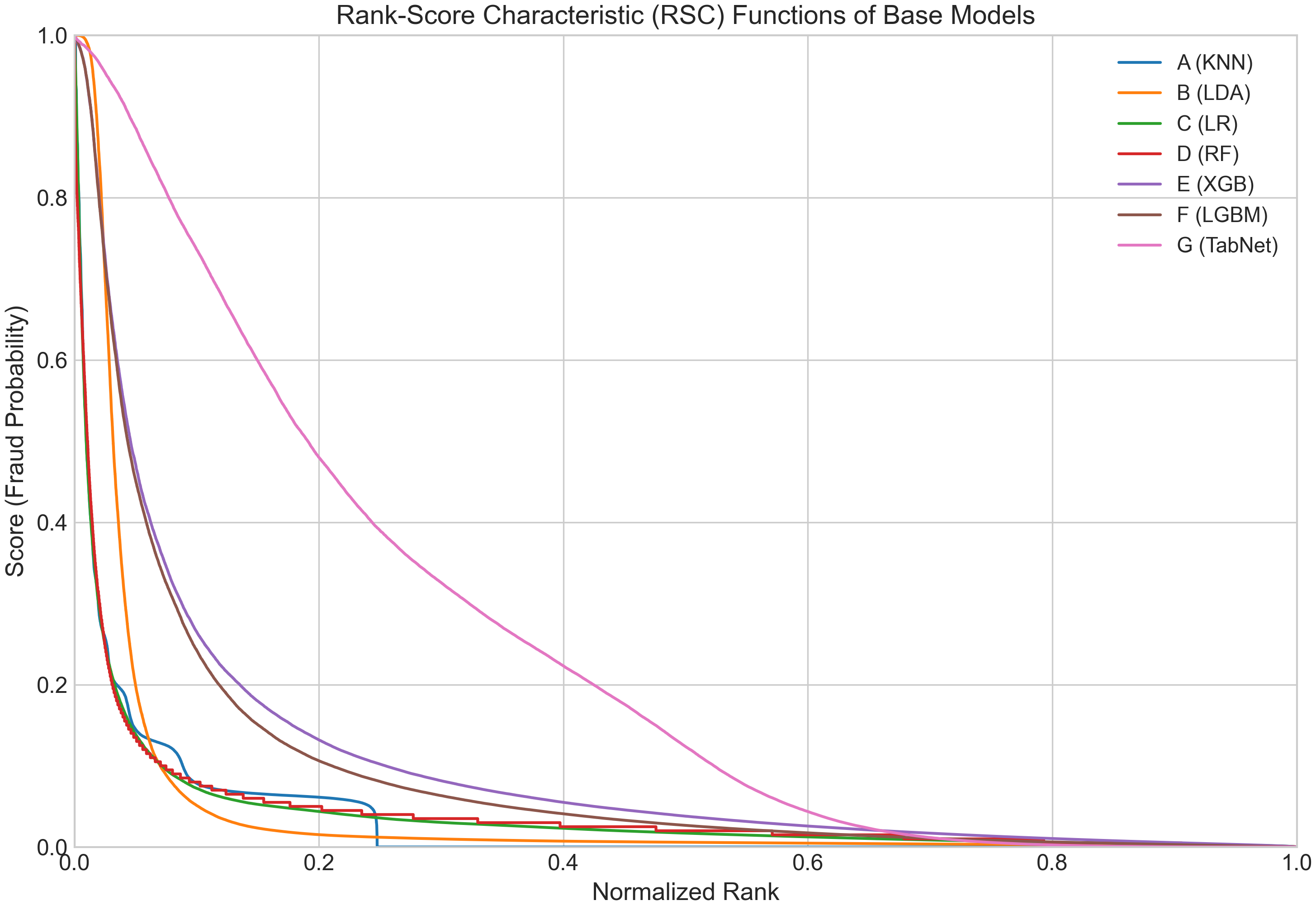}
\caption{Rank-Score Characteristic functions of the seven base models. XGBoost and LightGBM form sharp high-risk profiles; TabNet is visibly distinct; the linear models are flatter.}
\label{fig:rsc}
\end{figure}

\subsection{Diversity Structure}
The RSC diversity matrix in Table~\ref{tab:diversity} helps explain the CFA selection. XGBoost and LightGBM are very similar ($d=0.012$), which is expected for two boosted-tree methods, but they are also the strongest base learners. TabNet is the most diverse model, with distances from 0.144 to 0.211 from the others. That diversity does not make it part of the best subset, because its own ranking performance is weaker. In this benchmark, CFA balances model strength and complementarity rather than simply maximizing diversity.

\begin{table}[t]
\caption{Pairwise RSC diversity matrix on the validation set.}
\label{tab:diversity}
\centering
\scriptsize
\begin{tabular}{lrrrrrrr}
\toprule
 & A & B & C & D & E & F & G \\
\midrule
A & 0.000 & 0.027 & 0.014 & 0.016 & 0.070 & 0.058 & 0.211 \\
B & 0.027 & 0.000 & 0.028 & 0.031 & 0.059 & 0.047 & 0.199 \\
C & 0.014 & 0.028 & 0.000 & 0.004 & 0.063 & 0.051 & 0.205 \\
D & 0.016 & 0.031 & 0.004 & 0.000 & 0.061 & 0.049 & 0.203 \\
E & 0.070 & 0.059 & 0.063 & 0.061 & 0.000 & 0.012 & 0.144 \\
F & 0.058 & 0.047 & 0.051 & 0.049 & 0.012 & 0.000 & 0.154 \\
G & 0.211 & 0.199 & 0.205 & 0.203 & 0.144 & 0.154 & 0.000 \\
\bottomrule
\end{tabular}
\end{table}

\subsection{Validation Selection and Test Evaluation}
All 480 fusion cases are scored on the validation set. The same subset, DEF (Random Forest, XGBoost, and LightGBM), is selected for all three target metrics. AvgScore is marginally better for validation AUC-ROC, while WtScore is selected for AUPRC and F1. On the test set, DEF WtScore gives the highest AUPRC and F1 and essentially matches AvgScore on AUC-ROC.

\begin{table}[t]
\caption{Validation-selected CFA configurations evaluated on the test set.}
\label{tab:selection}
\centering
\scriptsize
\begin{tabular}{lllrrr}
\toprule
Selected by & Comb. & Method & Test AUC & Test AUPRC & Test F1 \\
\midrule
Val AUC & DEF & AvgScore & 0.9406 & 0.6694 & 0.6319 \\
Val AUPRC & DEF & WtScore & 0.9405 & 0.6699 & 0.6373 \\
Val F1 & DEF & WtScore & 0.9405 & 0.6699 & 0.6373 \\
\bottomrule
\end{tabular}
\end{table}

\subsection{Improvement over Single Models}
DEF WtScore improves over LightGBM by 0.0062 in AUC-ROC, 0.0219 in AUPRC, and 0.0280 in F1 (Table~\ref{tab:delta}). XGBoost is the stronger single-model F1 baseline, at 0.6126. CFA still improves over that value by 0.0247. The bootstrap intervals in Table~\ref{tab:bootstrap} exclude zero for all three metrics.

\begin{table}[t]
\caption{CFA versus the strongest single model on the test set.}
\label{tab:delta}
\centering
\scriptsize
\begin{tabular}{lrrr}
\toprule
Metric & LightGBM & DEF WtScore & Point $\Delta$ \\
\midrule
AUC-ROC & 0.9343 & 0.9405 & +0.0062 \\
AUPRC & 0.6480 & 0.6699 & +0.0219 \\
F1 & 0.6093 & 0.6373 & +0.0280 \\
\bottomrule
\end{tabular}
\end{table}

\begin{table}[t]
\caption{Bootstrap 95\% confidence intervals, DEF WtScore versus LightGBM.}
\label{tab:bootstrap}
\centering
\scriptsize
\begin{tabular}{lrrr}
\toprule
Metric & Mean $\Delta$ & Lower & Upper \\
\midrule
AUC-ROC & +0.0063 & +0.0048 & +0.0081 \\
AUPRC & +0.0213 & +0.0170 & +0.0257 \\
F1 & +0.0226 & +0.0159 & +0.0289 \\
\bottomrule
\end{tabular}
\end{table}

\subsection{Search Landscape}
\begin{figure*}[t]
\centering
\begin{minipage}[t]{0.48\textwidth}
\centering
\includegraphics[width=\linewidth]{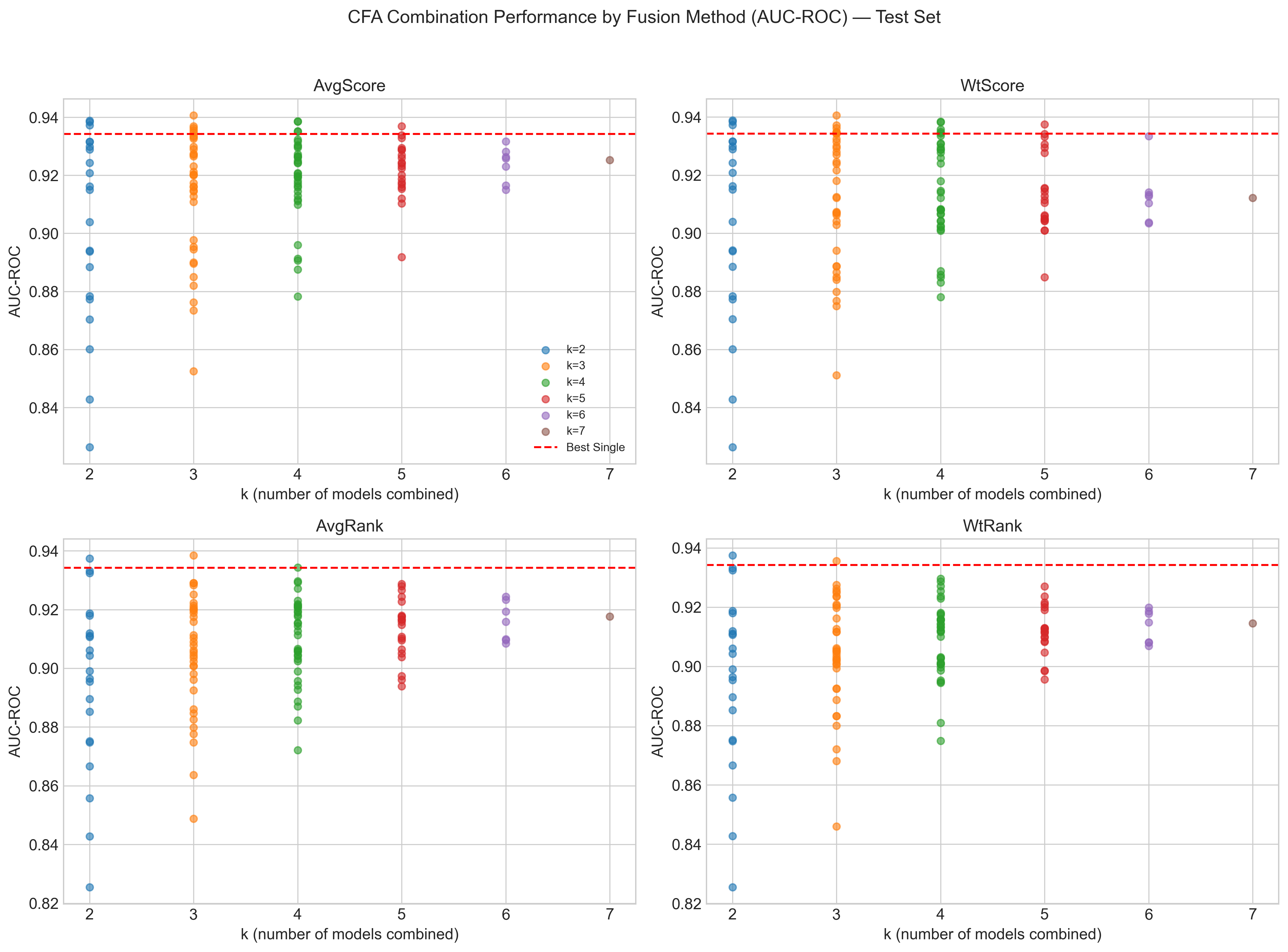}
\caption{Validation AUC-ROC across the 480 CFA cases by fusion rule and subset size. The dashed line marks the best single-model AUC-ROC. Score-based fusion remains strongest across the search.}
\label{fig:auc_landscape}
\end{minipage}\hfill
\begin{minipage}[t]{0.48\textwidth}
\centering
\includegraphics[width=\linewidth]{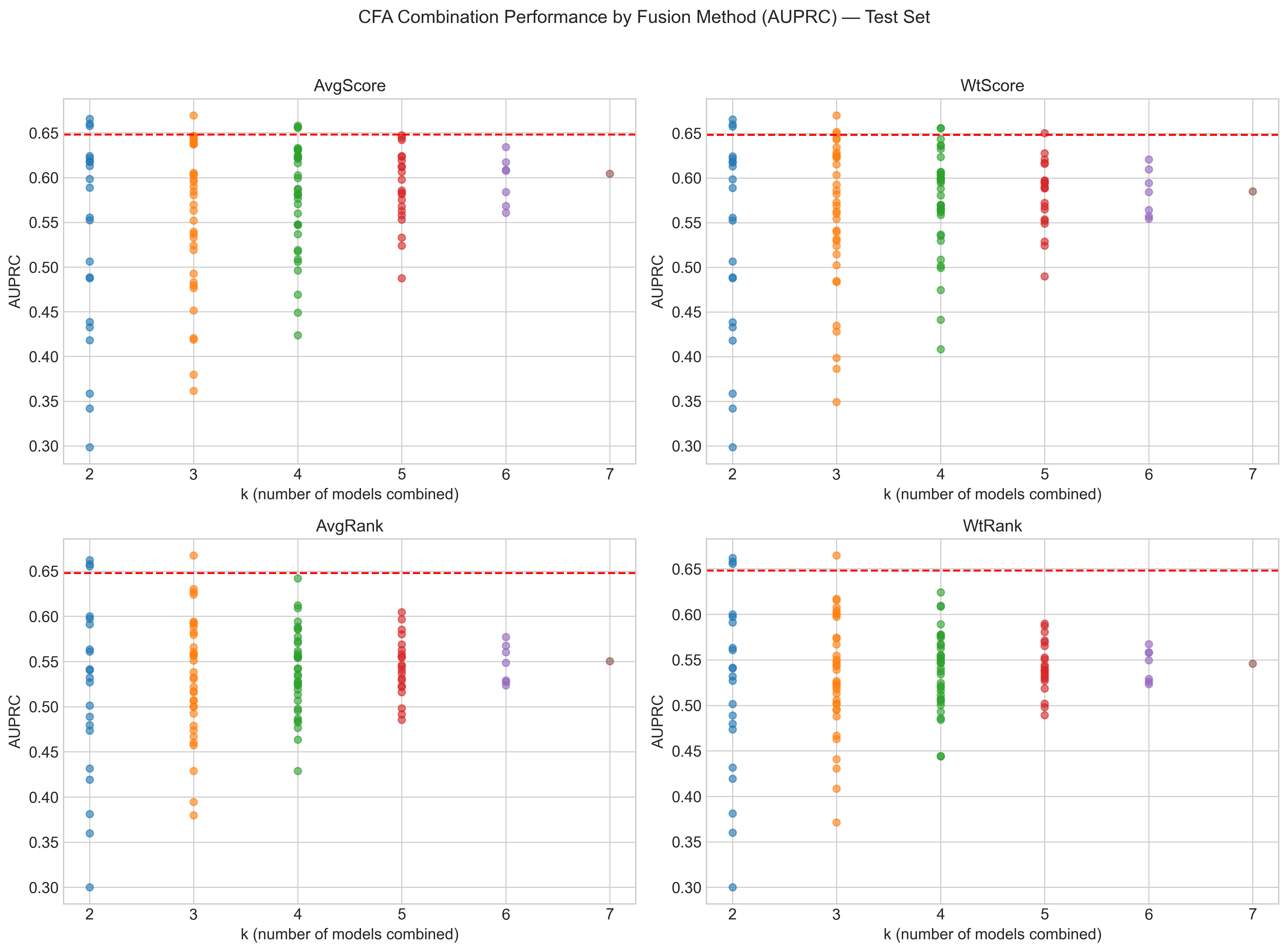}
\caption{Validation AUPRC across the same CFA cases. Score-based methods stay above the single-model baseline for compact subsets; rank-based methods show larger variance and weaker high-$k$ behavior.}
\label{fig:auprc_landscape}
\end{minipage}
\end{figure*}

Figures~\ref{fig:auc_landscape} and~\ref{fig:auprc_landscape} summarize the CFA search. The pattern is consistent: score-based fusion is stronger than rank-based fusion, and smaller subsets are more stable than larger ones. Rank fusion is not inherently weaker in CFA. Here, however, the model scores appear informative enough that replacing them with ranks removes useful separation.

\subsection{Comparison with Standard Ensembles}
Table~\ref{tab:standard} gives the most important comparison. CFA does not materially improve AUC-ROC over soft voting on the same selected subset, because CFA AvgScore on DEF and soft voting on DEF are the same operation. The added value is the structured validation-stage search that selects DEF from 120 candidate subsets, plus the small AUPRC and F1 gains from diversity-weighted score fusion. Stacking performs worse than the best single model, which is plausible when the base models are strong and highly correlated.

\begin{table}[t]
\caption{CFA versus standard ensemble baselines on the test set.}
\label{tab:standard}
\centering
\scriptsize
\begin{tabular}{lrrrrr}
\toprule
Method & AUC & AUPRC & F1 & Prec. & Rec. \\
\midrule
LightGBM & 0.9343 & 0.6480 & 0.6093 & 0.6640 & 0.5629 \\
Soft Voting DEF & 0.9406 & 0.6694 & 0.6319 & 0.7270 & 0.5588 \\
Stacking DEF & 0.9270 & 0.6287 & 0.6005 & 0.7361 & 0.5070 \\
CFA DEF WtScore & 0.9405 & 0.6699 & 0.6373 & 0.7139 & 0.5755 \\
\bottomrule
\end{tabular}
\end{table}

CFA DEF WtScore uses a lower decision threshold than soft voting (0.431 versus 0.505). It therefore trades about 1.3 percentage points of precision for about 1.7 points of recall, which produces the higher F1.

\subsection{Operational Threshold Metrics}
Because F1 describes only one operating point, we also report two metrics tied to fixed alert-budget decisions (Table~\ref{tab:operational}). At recall 0.50, where the system catches half of all fraud cases, CFA DEF WtScore reaches precision 0.8048, compared with 0.7669 for LightGBM. At precision 0.70, where 70\% of alerts are true fraud, CFA recovers 58.4\% of fraud, compared with 54.7\% for LightGBM. The gain is therefore not just an artifact of the F1 threshold.

\begin{table}[t]
\caption{Operational threshold metrics on the test set.}
\label{tab:operational}
\centering
\scriptsize
\setlength{\tabcolsep}{2pt}
\begin{tabular}{p{0.20\linewidth}rrrr}
\toprule
Metric & CFA DEF WtScore & LightGBM & XGBoost & Delta vs LGBM \\
\midrule
Precision at recall 0.50 & 0.8048 & 0.7669 & 0.7565 & +0.0379 \\
Recall at precision 0.70 & 0.5835 & 0.5472 & 0.5368 & +0.0363 \\
\bottomrule
\end{tabular}
\end{table}

\subsection{CTGAN Augmentation and Imbalance Robustness}
CTGAN is trained on the 12,398 training-set fraud samples for 150 epochs with batch size 500. Synthetic fraud rows are then generated until the training fraud rate is approximately balanced. In this configuration, augmentation hurts every retrained model and lowers CFA performance as well (Tables~\ref{tab:ctgan_models} and~\ref{tab:ctgan_cfa}). This result should be read narrowly. It reflects this numerical-sdtype setup, not a claim that CTGAN can never help fraud detection. It also warns against treating high-cardinality fields such as \texttt{card1}, \texttt{addr1}, and email domains as ordinary numeric variables during synthetic generation.

\begin{table}[t]
\caption{Effect of CTGAN augmentation on individual models.}
\label{tab:ctgan_models}
\centering
\scriptsize
\begin{tabular}{lrrr}
\toprule
Model & AUC-ROC & AUPRC & F1 \\
\midrule
D* RF + CTGAN & 0.8711 & 0.5159 & 0.5106 \\
E* XGB + CTGAN & 0.8976 & 0.5558 & 0.5530 \\
F* LGBM + CTGAN & 0.8988 & 0.5528 & 0.5506 \\
G* TabNet + CTGAN & 0.8827 & 0.5212 & 0.5391 \\
\bottomrule
\end{tabular}
\end{table}

\begin{table}[t]
\caption{Effect of CTGAN augmentation on CFA.}
\label{tab:ctgan_cfa}
\centering
\scriptsize
\begin{tabular}{lrrr}
\toprule
 & AUC-ROC & AUPRC & F1 \\
\midrule
Best CFA, original & 0.9406 & 0.6694 & 0.6319 \\
Best CFA, with CTGAN & 0.9103 & 0.6120 & 0.5765 \\
$\Delta$ & -0.0303 & -0.0574 & -0.0554 \\
\bottomrule
\end{tabular}
\end{table}

\begin{figure}[t]
\centering
\includegraphics[width=0.95\linewidth]{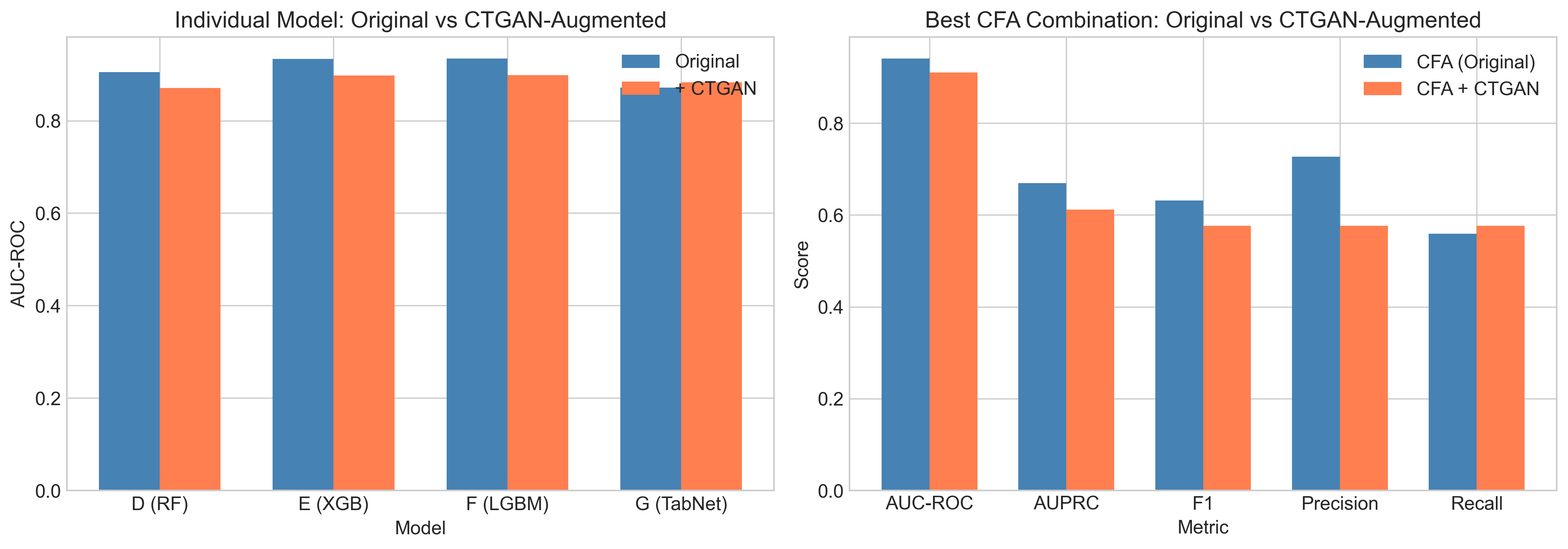}
\caption{CTGAN augmentation results. Individual model AUC-ROC decreases after augmentation, and the best CFA configuration also declines across AUC-ROC, AUPRC, and F1.}
\label{fig:ctgan}
\end{figure}

We also test robustness to class imbalance by undersampling the training set to fraud ratios of 0.5\%, 1\%, 3.5\%, 10\%, and 25\%. CFA WtScore matches or exceeds the best single model from 1\% upward. At 0.5\%, it underperforms, likely because too few fraud examples are available to estimate RSC diversity reliably.

\begin{figure}[t]
\centering
\includegraphics[width=0.95\linewidth]{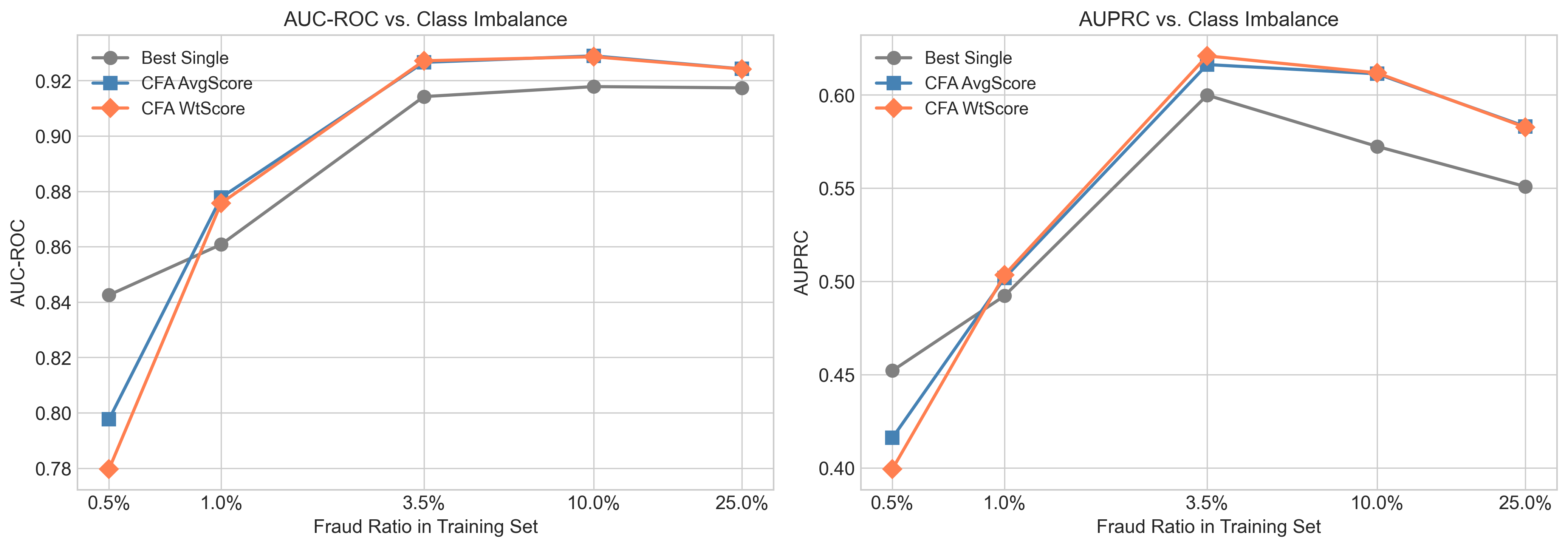}
\caption{Class-imbalance ablation. CFA tracks or exceeds the best single model once the training fraud ratio reaches 1\%; at 0.5\%, the diversity signal appears too noisy to help.}
\label{fig:imbalance}
\end{figure}

\section{Discussion}
The main result is straightforward. The best CFA configuration is not a large ensemble; it is Random Forest, XGBoost, and LightGBM combined by diversity-weighted score fusion. TabNet is the most diverse model by RSC distance, but it is not strong enough here to improve the fusion. The result suggests that diversity helps only when the added model makes useful errors, not merely different errors.

The gain is reliable but modest: about 0.6 percentage points in AUC-ROC and 2.2 points in AUPRC over LightGBM. A well-tuned LightGBM may be enough when broad ranking is the only goal. The F1 and operating-point gains are more relevant when alert volume and review cost matter. The CTGAN result is also practical: synthetic fraud rows can hurt when high-cardinality categorical fields are label-encoded and then treated as numeric.

The study has three main limitations. The split is stratified random rather than temporal, although \texttt{TransactionDT} would support time-based testing. The ensemble baselines are fair under the same protocol, but not exhaustively optimized over every subset. Finally, F1 is only one threshold metric; real fraud systems often optimize precision at a fixed alert budget, recall at a fixed false-positive rate, or expected cost.

\section{Conclusion}
The results point to a practical but limited finding. CFA with diversity-weighted score fusion improves over the strongest single model and over the tested voting and stacking baselines, with bootstrap intervals excluding zero. The best configuration, DEF WtScore, reaches AUC-ROC = 0.9405, AUPRC = 0.6699, and F1 = 0.6373. The selected subset is a small group of strong tree models, not the most diverse library. Thus, CFA is best viewed here as a validation-stage search method for compact, complementary subsets, not as evidence that larger ensembles or synthetic augmentation are automatically beneficial.

\balance


\begin{thebibliography}{23}

\bibitem{bhattacharyya2011} S. Bhattacharyya, S. Jha, K. Tharakunnel, and J. C. Westland, ``Data mining for credit card fraud: A comparative study,'' \textit{Decis. Support Syst.}, vol. 50, no. 3, pp. 602--613, 2011.

\bibitem{dalpozzolo2014} A. Dal Pozzolo, O. Caelen, Y.-A. Le Borgne, S. Waterschoot, and G. Bontempi, ``Learned lessons in credit card fraud detection from a practitioner perspective,'' \textit{Expert Syst. Appl.}, vol. 41, no. 10, pp. 4915--4928, 2014.

\bibitem{dalpozzolo2018} A. Dal Pozzolo, G. Boracchi, O. Caelen, C. Alippi, and G. Bontempi, ``Credit card fraud detection: A realistic modeling and a novel learning strategy,'' \textit{IEEE Trans. Neural Netw. Learn. Syst.}, vol. 29, no. 8, pp. 3784--3797, 2018.

\bibitem{carcillo2021} F. Carcillo, Y.-A. Le Borgne, O. Caelen, Y. Kessaci, F. Oble, and G. Bontempi, ``Combining unsupervised and supervised learning in credit card fraud detection,'' \textit{Inf. Sci.}, vol. 557, pp. 317--331, 2021.

\bibitem{he2009} H. He and E. A. Garcia, ``Learning from imbalanced data,'' \textit{IEEE Trans. Knowl. Data Eng.}, vol. 21, no. 9, pp. 1263--1284, Sep. 2009.

\bibitem{fawcett2006} T. Fawcett, ``An introduction to ROC analysis,'' \textit{Pattern Recognit. Lett.}, vol. 27, no. 8, pp. 861--874, Jun. 2006.

\bibitem{saito2015} T. Saito and M. Rehmsmeier, ``The precision-recall plot is more informative than the ROC plot when evaluating binary classifiers on imbalanced datasets,'' \textit{PLoS ONE}, vol. 10, no. 3, art. e0118432, Mar. 2015.

\bibitem{chen2016} T. Chen and C. Guestrin, ``XGBoost: A scalable tree boosting system,'' in \textit{Proc. ACM SIGKDD}, 2016, pp. 785--794.

\bibitem{ke2017} G. Ke \textit{et al.}, ``LightGBM: A highly efficient gradient boosting decision tree,'' in \textit{Advances in NeurIPS}, 2017, pp. 3146--3154.

\bibitem{elbazi2024} A. El Bazi, M. Chrayah, N. Aknin, and A. Bouzidi, ``Enhancing credit card fraud detection using a stacking model approach and hyperparameter optimization,'' \textit{Int. J. Adv. Comput. Sci. Appl.}, vol. 15, no. 10, 2024.

\bibitem{paldino2024} G. M. Paldino \textit{et al.}, ``The role of diversity and ensemble learning in credit card fraud detection,'' \textit{Adv. Data Anal. Classif.}, vol. 18, no. 1, pp. 193--217, 2024.

\bibitem{arik2021} S. O. Arik and T. Pfister, ``TabNet: Attentive interpretable tabular learning,'' in \textit{Proc. AAAI}, vol. 35, no. 8, 2021, pp. 6679--6687.

\bibitem{jurgovsky2018} J. Jurgovsky \textit{et al.}, ``Sequence classification for credit-card fraud detection,'' \textit{Expert Syst. Appl.}, vol. 100, pp. 234--245, 2018.

\bibitem{chawla2002} N. V. Chawla, K. W. Bowyer, L. O. Hall, and W. P. Kegelmeyer, ``SMOTE: Synthetic minority over-sampling technique,'' \textit{J. Artif. Intell. Res.}, vol. 16, pp. 321--357, 2002.

\bibitem{dalpozzolo2015} A. Dal Pozzolo, O. Caelen, R. A. Johnson, and G. Bontempi, ``Calibrating probability with undersampling for unbalanced classification,'' in \textit{Proc. IEEE SSCI}, 2015, pp. 159--166.

\bibitem{xu2019} L. Xu, M. Skoularidou, A. Cuesta-Infante, and K. Veeramachaneni, ``Modeling tabular data using conditional GAN,'' in \textit{Advances in NeurIPS}, 2019, pp. 7335--7345.

\bibitem{fiore2019} U. Fiore, A. De Santis, F. Perla, P. Zanetti, and F. Palmieri, ``Using GANs for improving classification effectiveness in credit card fraud detection,'' \textit{Inf. Sci.}, vol. 479, pp. 448--455, 2019.

\bibitem{hsu2006} D. F. Hsu, Y.-S. Chung, and B. S. Kristal, ``Combinatorial fusion analysis: Methods and practices of combining multiple scoring systems,'' in \textit{Advanced Data Mining Technologies in Bioinformatics}. IGI Global, 2006, pp. 32--62.

\bibitem{hsu2005} D. F. Hsu and I. Taksa, ``Comparing rank and score combination methods for data fusion in information retrieval,'' \textit{Inf. Retr.}, vol. 8, no. 3, pp. 449--480, 2005.

\bibitem{hsu2010} D. F. Hsu, B. S. Kristal, and C. Schweikert, ``Rank-score characteristics (RSC) function and cognitive diversity,'' in \textit{Brain Informatics, LNAI 6334}. Springer, 2010, pp. 42--54.

\bibitem{schweikert2012} C. Schweikert, S. Brown, Z. Tang, P. R. Smith, and D. F. Hsu, ``Combining multiple ChIP-seq peak detection systems using combinatorial fusion,'' \textit{BMC Genomics}, vol. 13, suppl. 8, art. S12, 2012.

\bibitem{lin2024} C. Lin, C. Schweikert, and D. F. Hsu, ``Identifying distributed denial of service attacks through multi-model deep learning fusion and combinatorial analysis,'' \textit{J. Netw. Syst. Manag.}, vol. 32, art. 89, 2024.

\bibitem{hsu2024} D. F. Hsu, B. S. Kristal, and C. Schweikert, ``Combinatorial fusion analysis,'' \textit{Computer}, vol. 57, no. 9, pp. 96--100, Sep. 2024.

\end{thebibliography}
\end{document}